\RequirePackage{ifthen}
\newboolean{MPA}
\setboolean{MPA}{false}

\ifthenelse {\boolean{MPA}}
{
\documentclass[smallextended]{svjour3}
\smartqed
} {
\documentclass[11pt]{article}
\usepackage[margin=1.3in]{geometry}
}

\usepackage{hyperref}

\usepackage{graphicx}
\usepackage{amsmath}
\usepackage{amssymb}
\usepackage{color}

\usepackage{enumitem} 
\setlist{nosep}

\newcommand{\R}{\mathbb R}
\newcommand{\Z}{\mathbb Z}
\newcommand{\Q}{\mathbb Q}
\newcommand{\transp}{\mathsf T}

\newcommand{\din}{\delta^{\text{in}}}
\newcommand{\dout}{\delta^{\text{out}}}

\DeclareMathOperator{\trace}{trace}
\DeclareMathOperator{\rk}{rank}
\DeclareMathOperator{\ext}{ext}
\DeclareMathOperator{\rst}{RST}
\DeclareMathOperator{\crc}{CRC}
\DeclareMathOperator{\supp}{supp}

\newcommand{\C}{\mathcal C}
\newcommand{\E}{\mathcal E}
\renewcommand{\H}{\mathcal H}
\renewcommand{\S}{\mathcal S}
\newcommand{\X}{\mathcal X}

\newcommand{\abs}[1]{\lvert#1\rvert}
\newcommand{\norm}[1]{\lVert#1\rVert_2}
\newcommand{\Fnorm}[1]{\lVert#1\rVert_F}
\newcommand{\pare}[1]{\left(#1\right)}
\newcommand{\bra}[1]{\{#1\}}

\newcommand{\st}{\ : \ }

\ifthenelse {\boolean{MPA}}
{

\newenvironment{prf}[1][]
{\begin{proof}}
{\qed \end{proof}}

\newcounter{claim}
\renewenvironment{claim}
{\refstepcounter{claim} \begin{trivlist} \item[] {\em Claim~\theclaim}\space \itshape}
{\end{trivlist}}
\newenvironment{cpf}
{\begin{trivlist} \item[] {\em Proof of claim}\space}
{$\hfill\diamond$ \end{trivlist}}

\newenvironment{cpfnodiamond}
{\begin{trivlist} \item[] {\em Proof of claim}\space}
{\end{trivlist}}

\journalname{Mathematical Programming A}

} {

\usepackage{amsthm}
\newtheorem{theorem}{Theorem}
\newtheorem{proposition}{Proposition}
\newtheorem{lemma}{Lemma}

\newtheorem{claim}{Claim}

\newenvironment{prf}[1][]
{\begin{proof}}
{\end{proof}}

\newenvironment{cpf}
{\begin{trivlist} \item[] {\em Proof of claim. }}
{$\hfill\diamond$ \end{trivlist}}

\newenvironment{cpfnodiamond}
{\begin{trivlist} \item[] {\em Proof of claim. }}
{\end{trivlist}}

}

\begin{document}

\title{Sparse PCA on fixed-rank matrices \thanks{This work is supported by ONR grant N00014-19-1-2322. Any opinions, findings, and conclusions or recommendations expressed in this material are those of the authors and do not necessarily reflect the views of the Office of Naval Research.}}

\ifthenelse {\boolean{MPA}}
{
\titlerunning{Sparse PCA on fixed-rank matrices}

\author{Alberto Del Pia}
\institute{Alberto~Del~Pia \at
              Department of Industrial and Systems Engineering 
              \& Wisconsin Institute for Discovery \\
              University of Wisconsin-Madison, Madison, WI, USA \\
              \email{delpia@wisc.edu}}
}
{
\author{Alberto Del Pia
\thanks{Department of Industrial and Systems Engineering \& Wisconsin Institute for Discovery,
             University of Wisconsin-Madison, Madison, WI, USA.
             E-mail: {\tt delpia@wisc.edu}.}}
}

\date{\today}

\maketitle

\begin{abstract}
Sparse PCA is the optimization problem obtained from PCA by adding a sparsity constraint on the principal components.
Sparse PCA is NP-hard and hard to approximate even in the single-component case.
In this paper we settle the computational complexity of sparse PCA with respect to the rank of the covariance matrix.
We show that, if the rank of the covariance matrix is a fixed value, then there is an algorithm that solves sparse PCA to global optimality, whose running time is polynomial in the number of features.
We also prove a  similar result for the version of sparse PCA which requires the principal components to have disjoint supports.
\ifthenelse {\boolean{MPA}}
{
\keywords{principal component analysis \and sparsity \and polynomial-time algorithm \and global optimum \and constant-rank quadratic function}
\subclass{MSC 90C20 \and 90C26 \and 90C60 \and 68Q25}
} {}
\end{abstract}

\ifthenelse {\boolean{MPA}}
{}{
\emph{Key words:} principal component analysis; sparsity; polynomial-time algorithm; global optimum; constant-rank quadratic function
}

\section{Introduction}

\emph{Principal component analysis} is one of the oldest and most popular dimensionality reduction techniques and it is used in a wide array of scientific disciplines.
In principal component analysis, we are given a positive integer $d$ and an $n \times m$ \emph{data matrix} $Q$, where each column represents an independent sample from data population, and each row gives a particular kind of feature.
Our task is to find $d$ linear combinations of the $n$ features, called \emph{principal components}, that correspond to directions of maximal variance in the data. 
The $d$ principal components typically explain most of the variance present in the data, even if the number $d$ is chosen to be much lower than the number of features $n$ in the original dataset.
Typically, principal component analysis is formulated in terms of the \emph{covariance matrix}, which is the $n \times n$ positive semidefinite matrix $K := 1/m \cdot (Q-E[Q]) (Q-E[Q])^\transp$.
Formally, in principal component analysis we are given an $n \times n$ positive semidefinite matrix $K$, a positive integer $d$ smaller than $n$, and we seek an optimal solution to the optimization problem
\begin{align}
\label{prob pca}
\tag{PCA}
\max_{X \in \R^{n \times d}, \  X^\transp X = I_d} \
\trace(X^\transp K X),
\end{align}
where $I_d$ denotes the $d \times d$ identity matrix.
The $d$ principal components correspond to the $d$ columns of an optimal solution $X$.
It is well-known that \ref{prob pca} can be efficiently solved. 
In fact, an optimal solution is the matrix $X$ whose columns are the $d$ eigenvectors of $K$ corresponding to the largest $d$ eigenvalues.
This optimal solution to \ref{prob pca} can be found in $O(n^3)$ time by computing an eigenvalue decomposition of $K$.
We refer the reader to \cite{ShaBenMLbook} for an introduction to principal component analysis.

\subsection{Sparse PCA}
\label{sec intro spca}

A potential disadvantage of \ref{prob pca} is that the principal components are usually linear combinations of all features. 
This often makes the derived principal components difficult to interpret.
\emph{Sparse principal component analysis} overcomes this disadvantage by requiring the principal components to be linear combinations of just a few features.
A direct consequence is that sparse principal component analysis generally provides higher data interpretability as well as better generalization error \cite{CadJol95,JolTreUdd03,HasTibWai15,ZouHasTib06,BouDriMag11}.
A natural formulation of sparse principal component analysis is obtained by adding to \ref{prob pca} a sparsity constraint on the principal components.
Formally, in sparse principal component analysis we are given an $n \times n$ positive semidefinite matrix $K$, positive integers $d,s$ smaller than $n$, and we seek an optimal solution to the optimization problem
\begin{align}
\label{prob spca}
\tag{SPCA}
\max_{\substack{X \in \R^{n \times d} \\ X^\transp X = I_d, \ \abs{\supp(X)} \le s}} \ 
\trace(X^\transp K X),
\end{align}
where $\supp(X)$ denotes the index set of the nonzero rows of the matrix $X$.
Throughout this paper we will often discuss the special cases of \ref{prob pca} and \ref{prob spca} with $d=1$.
We refer to these cases, where we only seek one principal component, as the \emph{single-component} cases.

\ref{prob spca} is NP-hard and hard to approximate~\cite{ChaPapRub16,Mag17} even in the single component case.
Successful approaches for \ref{prob spca} include replacing the $\ell_0$-norm constraint with an $\ell_1$-norm constraint or $\ell_1$ penalty \cite{JolTreUdd03,ZouHasTib06,VuLei12},
branch-and bound \cite{MogWeiAvi06NIPS,BerBer19},
semidefinite programming \cite{dAsElGJorLan07,dAsBacGha08,ZhadAsGha12,dAsBacGha14},
and convex integer programming \cite{DeyMazWan18}.
A number of other specialized algorithms have been proposed in, e.g., \cite{SigBuh08,HeMonPar10,JouNesRicSep10,BouDriMag11,AstPapKar11,YuaZha13,PapDimKor13}.
Only few of these papers directly deal with the general version of \ref{prob spca} as defined in this paper \cite{BouDriMag11}. 
In fact, most known algorithms are based on an iterative approach where the principal components are estimated in a one-at-a-time fashion with some sort of deflation step between iterations \cite{Mac09}.

The main challenge in solving \ref{prob spca} to global optimality lies in identifying an optimal support of \ref{prob spca} among all the $\binom{n}{s}$ index sets of cardinality $s$, where  an \emph{optimal support} of \ref{prob spca} is defined as an index set $S^* \subseteq \{1,\dots,n\}$ of cardinality $s$ such that $\supp(X^*) \subseteq S^*$ for an optimal solution $X^*$ to \ref{prob spca}.
Asteris et al.~\cite{AstPapKar14} show that, in the single-component case, it is possible to design an algorithm that identifies $O(n^r)$ candidate supports in $O(n^{r+1})$ time, where $r$ denotes the rank of the matrix $K$, among which lies an optimal support.
Therefore, if one considers matrices $K$ whose rank $r$ is a fixed value, both the number of candidate supports constructed and the running time of the algorithm are polynomial in $n$.
In this paper, we confirm that fixing the rank $r$ of $K$ is key in solving \ref{prob spca} in polynomial time, and not just in the single-component case, but for any number $d$ of principal components.
Next, we formally state our first main result.

\begin{theorem}
\label{th spca}
There is an algorithm that finds an optimal solution to \ref{prob spca} in time 
$$
O\left(n^{\min\{d,r\}(r^2+r)}(\min\{d,r\}n r^2 + n \log n)\right),
$$
where $r$ denotes the rank of the input matrix $K$.
In particular, the algorithm constructs $O(n^{\min\{d,r\}(r^2+r)})$ candidate supports among which lies an optimal support.
\end{theorem}

If the rank $r$ of $K$ is a fixed value, then both the number of candidate supports constructed and the running time of the algorithm are polynomial in $n$.
Theorem~\ref{th spca} constitutes the first polynomial-time algorithm for \ref{prob spca}, for any fixed value of $r$.
We remark that the running time exponential dependence on $r$ is expected, since \ref{prob spca} is NP-hard in its full generality.
The proof of Theorem~\ref{th spca} is given in Section~\ref{sec identical}.

\subsection{Sparse PCA with disjoint supports}
\label{sec intro spca-ds}

In this paper, we study also \emph{sparse principal component analysis with disjoint supports,} which is a different version of sparse principal component analysis which has been considered in the literature (see, e.g., \cite{AstPapKyrDim15}). 
Also in this model each principal component is a linear combination of at most $s$ features, but here no feature can be used by two different principal components.
Given a matrix $X$, we denote by $x_i$ its $i$th column.
Furthermore, for a nonnegative integer $d$, we let $[d] := \{1,\dots,d\}$.
With this notation, we can denote by $\X$ the set of feasible matrices
\begin{align*}
\X := \{X \in \R^{n \times d} \st & \abs{\supp(x_i)} \le s, \ \norm{x_i} = 1, \ \forall i \in [d], \\
& \supp(x_i) \cap \supp(x_{i'}) = \emptyset, \ \forall i \neq i' \in [d]\},
\end{align*}
Formally, in sparse principal component analysis with disjoint supports we are given an $n \times n$ positive semidefinite matrix $K$, positive integers $d,s$ smaller than $n$, and we seek an optimal solution to the optimization problem
\begin{align}
\label{prob spca-ds}
\tag{SPCA-DS}
& \max_{X \in \X} \ 
\trace(X^\transp K X).
\end{align}
We remark that single-component \ref{prob spca} is also a special case of \ref{prob spca-ds}, obtained by setting $d=1$.
Therefore, also \ref{prob spca-ds} is NP-hard and hard to approximate.

Similarly to \ref{prob spca}, the main difficulty in \ref{prob spca-ds} consists in finding an optimal support of \ref{prob spca-ds} among all the $O(n^{ds})$ families of $d$ index sets of cardinality at most $s$, where an \emph{optimal support} of \ref{prob spca-ds} is defined as a family of index sets $\bra{S_i^*}_{i \in [d]}$ with $S_i^* \subseteq [n]$, $|S_i^*| \le s$, $\forall i \in [d]$, $S_i^* \cap S_{i'}^* = \emptyset$, $\forall i \neq i' \in [d]$, and such that $\supp(x_i^*) \subseteq S_i^*$, $\forall i \in [d]$, for an optimal solution $X^*$ to \ref{prob spca-ds}.
Our second main result, stated below, implies that we can construct $O((dn)^{d^2(r^2+r)/2})$ candidate supports, among which lies an optimal one.

\begin{theorem}
\label{th spca-ds}
There is an algorithm that finds an optimal solution to \ref{prob spca-ds} in time
$$
O\left( 
(dn)^{d^2(r^2+r)/2} (dn r^2 + d^3 n^5 \log n)
\right),
$$
where $r$ denotes the rank of the input matrix $K$. 
In particular, the algorithm constructs $O((dn)^{d^2(r^2+r)/2})$ candidate supports, among which lies an optimal support.
\end{theorem}

If $r$ and $d$ are fixed values, then both the number of candidate supports constructed and the running time of the algorithm are polynomial in $n$.
Theorem~\ref{th spca-ds} then yields the first polynomial-time algorithm for \ref{prob spca-ds}, for any fixed values of $r$ and $d$.
To the best of our knowledge, the only other algorithm for \ref{prob spca-ds} with theoretical guarantees is given in~\cite{AstPapKyrDim15}, where the authors propose an algorithm that finds an $\epsilon$-approximate solution with running time polynomial in $n$ and $1/\epsilon$, provided that $r$ and $d$ are fixed.
The proof of Theorem~\ref{th spca-ds} can be found in Section~\ref{sec disjoint}.

\subsection{Techniques}
\label{sec techniques}

We briefly explain the main techniques used in our two algorithms.
To simplify the exposition, we assume that $r$ is a fixed value in \ref{prob spca}, and that both $r$ and $d$ are fixed in \ref{prob spca-ds}.

The first technique that we introduce is a dimensionality reduction approach which
allows us, in both problems, to replace our original matrix $X$ of variables with a new matrix $Y$ of variables which has the advantage of having only a fixed number of entries.
This approach 
can be seen as a multi-component generalization of the auxiliary unit vector technique
\cite{Mac94,Swe01,MotKriAna07,KarPad07,KarLia10,AstPapKar14}, and has strong connections with procedures used in principal component analysis when the original dimensionality $n$ of the data is much larger than the number of data vectors (see Section 23.1.1 in \cite{ShaBenMLbook}).

The next technique is a tool from discrete geometry known as the \emph{hyperplane arrangement theorem}.
A set $\H$ of $p$ hyperplanes in a $q$-dimensional Euclidean space determines a partition of the space called the \emph{arrangement of $\H$.} 
The hyperplane arrangement theorem states that this arrangement consists of $O(p^q)$ full-dimensional polyhedra and can be constructed in time $O(p^q)$.
For more details, we refer the reader to \cite{EdeOroSei86}, and in particular to Theorem~3.3 therein.
In both our algorithms, this theorem is employed to partition an extended version of the space of variables $Y$ in a polynomial number of polyhedra.
Each one will correspond to a candidate support that we construct, and at least one of them will be optimal to the problem.

Finally, in the proof of Theorem~\ref{th spca-ds}, we reduce a restricted version of \ref{prob spca-ds} to a maximum-profit integer circulation problem. 
This allows us to make use of the optimality conditions for this problem and of the strongly polynomial-time algorithm by Goldberg and Tarjan~\cite{GolTar88,GolTar89}.
First, the optimality conditions are exploited to obtain the arrangement discussed above.
Next, for each polyhedron in the arrangement, we select a vector in its interior and apply Goldberg and Tarjan's algorithm to the corresponding instance.
The output of the algorithm allows us to obtain the candidate support $\bra{S_i}_{i \in [d]}$ associated with the polyhedron.

\subsection{Computational complexity and practical applicability of our algorithms}

We remark that we do not expect that a direct implementation of our algorithms will lead to practical algorithms for solving \ref{prob spca} and \ref{prob spca-ds}.
Rather, our results demonstrate that these problems are efficiently solvable from a theoretical point of view in the settings considered. 
This is important, because once a problem is shown to be efficiently solvable, usually practical algorithms follow (see, e.g., \cite{BerTsi97}).



We remark that our analysis of the algorithms can be improved in several ways to obtain marginally better running times.
For example, the hyperplane arrangement theorem is always used with a set $\H$ of $p$ hyperplanes \emph{that pass through the origin} in a $q$-dimensional Euclidean space.
In this special case, it is known that the arrangement consists of $O((p-1)^{q-1})$ full-dimensional polyhedra and can be constructed in time $O((p-1)^{q-1})$.

\section{A useful lemma}

Before proving our main results, we present a lemma that uses standard eigenvalue arguments.
This lemma plays a crucial role in the dimensionality reduction performed by both our algorithms. 
In particular, it implies that the optimal value of a \ref{prob pca} problem with an input matrix of fixed rank can be obtained by solving a different \ref{prob pca} problem with an input matrix of fixed dimensions.
In this paper, we denote by $\Fnorm{\cdot}$ the Frobenius norm.

\begin{lemma}
\label{lem xy}
Let $M$ be an $s \times r$ matrix, let $d$ be a positive integer, and let $d' := \min\{d,r\}$.
Then
\begin{align*}
\max_{\substack{X \in \R^{s \times d} \\ X^\transp X = I_d}} \ 
\Fnorm{M^\transp X}^2
=
\max_{\substack{X \in \R^{s \times d'} \\ X^\transp X = I_{d'}}} \ 
\Fnorm{M^\transp X}^2
=
\max_{\substack{Y \in \R^{r \times d'} \\ Y^\transp Y = I_{d'}}} \ 
\Fnorm{M Y}^2.
\end{align*}
\end{lemma}

\begin{prf}
Denote by $\lambda_j$, for $j \in [s]$, the eigenvalues of the $s \times s$ positive semidefinite matrix $MM^\transp$, and assume without loss of generality that $\lambda_1 \ge \lambda_2 \ge \cdots \ge \lambda_s \ge 0$.
Then
\begin{align}
\label{eq lem 1 a}
\max_{\substack{X \in \R^{s \times d} \\ X^\transp X = I_d}} \ 
\Fnorm{M^\transp X}^2
=
\max_{\substack{X \in \R^{s \times d} \\ X^\transp X = I_d}} \ 
\trace(X^\transp M M^\transp X)
=
\sum_{j=1}^d \lambda_j,
\end{align}
where in the first equality we used the definition of Frobenius norm and the second is well known (see, e.g., \cite{ShaBenMLbook}).

Symmetrically, we obtain 
\begin{align}
\label{eq lem 1 b}
\max_{\substack{X \in \R^{s \times d'} \\ X^\transp X = I_{d'}}} \ 
\Fnorm{M^\transp X}^2
=
\max_{\substack{X \in \R^{s \times d'} \\ X^\transp X = I_{d'}}} \ 
\trace(X^\transp M M^\transp X)
=
\sum_{j=1}^{d'} \lambda_j.
\end{align}
Since the nonzero eigenvalues of $M M^\transp$ are at most $\rk(MM^\transp) = \rk(M) \le r$, we have
$\sum_{j=1}^d \lambda_j = \sum_{j=1}^{d'} \lambda_j.$ 
Thus \eqref{eq lem 1 a} and \eqref{eq lem 1 b} coincide and we have shown the first equality in the statement of the lemma.

Denote by $\mu_k$, for $k \in [r]$, the eigenvalues of the $r \times r$ positive semidefinite matrix $M^\transp M$, and assume without loss of generality that $\mu_1 \ge \mu_2 \ge \cdots \ge \mu_r \ge 0$.
Similarly to our previous derivations, we have 
\begin{align}
\label{eq lem 1 c}
\max_{\substack{Y \in \R^{r \times d'} \\ Y^\transp Y = I_{d'}}} \ 
\Fnorm{M Y}^2
=
\max_{\substack{Y \in \R^{r \times d'} \\ Y^\transp Y = I_{d'}}} \ 
\trace(Y^\transp M^\transp M Y)
=
\sum_{k=1}^{d'} \mu_k.
\end{align}
Since the nonzero eigenvalues of $M M^\transp$ and $M^\transp M$ are the same, we have
$
\sum_{j=1}^{d'} \lambda_j = \sum_{k=1}^{d'} \mu_k.
$
Thus \eqref{eq lem 1 b} and \eqref{eq lem 1 c} coincide and we have shown the second equality in the statement of the lemma.
\end{prf}

\section{Proof of Theorem~\ref{th spca}}
\label{sec identical}

Consider \ref{prob spca} where the input matrix $K \in \R^{n \times n}$ has rank $r$.
Since the matrix $K$ is positive semidefinite, it is well known that we can compute an $n \times r$ matrix $R$ such that $K = RR^\transp$ in $O(n^3)$ time, for instance using the Cholesky decomposition with complete pivoting.
Using the definition of Frobenius norm, \ref{prob spca} takes the form
\begin{align}
\label{prob spca Frob}
\max_{\substack{X \in \R^{n \times d} \\ X^\transp X = I_d, \ \abs{\supp(X)} \le s}} \ 
\Fnorm{R^\transp X}^2.
\end{align}

We introduce some notation that will be used in this proof.
For $j \in [n]$, we denote by $R_j$ the $j$th row of $R$.
Similarly, for $S \subseteq [n]$, $R_S$ denotes the $|S| \times r$ submatrix of $R$ containing only the rows indexed by $S$.
We also denote by $d' := \min\{d,r\}$.

As discussed in Section~\ref{sec intro spca}, the main difficulty in solving Problem~\eqref{prob spca Frob} consists in finding an optimal support $S^*$ of Problem~\eqref{prob spca Frob}.
In fact, once $S^*$ is determined, an optimal solution $X^*$ to Problem~\eqref{prob spca Frob} can be obtained by setting to zero the rows of $X^*$ with indices not in $S^*$, while the other rows of $X^*$ can be obtained by solving the optimization problem 
\begin{align}
\label{prob spca final}
\max_{\substack{X \in \R^{s \times d} \\ X^\transp X = I_d}} \ 
\Fnorm{R_{S^*}^\transp X}^2.
\end{align}
This is a \ref{prob pca} problem with an $s \times s$ input matrix. 
In particular, the input matrix $R_{S^*} R_{S^*}^\transp$ can be constructed in $O(s^2r)$ time and an optimal solution can be found in $O(s^3)$ time.
Based on this discussion, in the remainder of the proof it suffices to find an optimal support $S^*$ of Problem~\eqref{prob spca Frob}.

The next claim uses Lemma~\ref{lem xy} to replace our matrix of variables $X \in \R^{n \times d}$ in Problem~\eqref{prob spca Frob} with an $r \times d'$ matrix of variables, that we denote by $Y$.
In the claim we consider the following two optimization problems:
\begin{align}
\label{prob spca X}
\max_{\substack{S \subseteq [n] \\ |S| = s}} \ \ 
\max_{\substack{X \in \R^{s \times d} \\ X^\transp X = I_d}} \ 
\Fnorm{(R_S)^\transp X}^2,
\end{align}
\begin{align}
\label{prob spca Y}
\max_{\substack{S \subseteq [n] \\ |S| = s}} \ \ 
\max_{\substack{Y \in \R^{r \times d'} \\ Y^\transp Y = I_{d'}}} \ 
\Fnorm{R_S Y}^2.
\end{align}
We say that $S^*$ is an \emph{optimal support} of Problem~\eqref{prob spca X} if there exists $X^*$ such that $(S^*,X^*)$ is an optimal solution to Problem~\eqref{prob spca X}.
Similarly, we say that $S^*$ is an \emph{optimal support} of Problem~\eqref{prob spca Y} if there exists $Y^*$ such that $(S^*,Y^*)$ is an optimal solution to Problem~\eqref{prob spca Y}.

\begin{claim}
\label{claim identical reduction}
The optimal supports of Problems~\eqref{prob spca Frob}, \eqref{prob spca X}, \eqref{prob spca Y} coincide.
\end{claim}

\begin{cpf}
Lemma~\ref{lem xy}, applied with $M := R_S$, implies that the optimal supports of Problems~\eqref{prob spca X} and \eqref{prob spca Y} coincide. 
Thus we only need to show that the optimal supports of Problems~\eqref{prob spca Frob} and \eqref{prob spca X} coincide.
To do so, it suffices to prove the following two statements:
(i) For every feasible solution $(S,X)$ to Problem~\eqref{prob spca X} with objective function value $\gamma$, there is a feasible solution $\tilde X$ to Problem~\eqref{prob spca Frob} with objective function value $\gamma$ such that $\supp(\tilde X) \subseteq S$;
(ii) For every feasible solution $\tilde X$ to Problem~\eqref{prob spca Frob} with objective function value $\gamma$, there is a feasible solution $(S,X)$ to Problem~\eqref{prob spca X} with objective function value $\gamma$ such that $\supp(\tilde X) \subseteq S$.

(i). Let $(S,X)$ be a feasible solution to Problem~\eqref{prob spca X} with objective function value $\gamma$.
Let $\tilde X \in \R^{n \times d}$ be obtained from $X$ by adding zero rows corresponding to the indices not in $S$. 
Then $\tilde X$ is a feasible solution to Problem~\eqref{prob spca Frob} with objective function value $\gamma$ such that $\supp(\tilde X) \subseteq S$.

(ii).
Let $\tilde X$ be a feasible solution to Problem~\eqref{prob spca Frob} with objective function value $\gamma$.
Let $S$ be a subset of $[n]$ of cardinality $s$ containing $\supp(\tilde X)$, and let $X$ be obtained from $\tilde X$ by dropping the (zero) rows with indices not in $S$.
Then $(S,X)$ is a feasible solution to Problem~\eqref{prob spca X} with objective function value $\gamma$ such that $\supp(\tilde X) \subseteq S$.
\end{cpf}

Due to Claim \ref{claim identical reduction}, in the rest of the proof our goal will be finding an optimal support of Problem~\eqref{prob spca Y}.
Next, we define a restricted version of Problem~\eqref{prob spca Y}, where we fix the matrix of variables $Y \in \R^{r \times d'}$:
\begin{align*}
\max_{\substack{S \subseteq [n] \\ |S| = s}} \ 
\Fnorm{R_S Y}^2.
\end{align*}
We denote this restricted problem by $\rst(Y)$.
The next claim gives a simple characterization of the optimal solutions to Problem~$\rst(Y)$.

\begin{claim}
\label{claim multi aux}
Let $Y \in \R^{r \times d'}$ be given. 
Then $S^*$ is an optimal solution to Problem~$\rst(Y)$ if and only if $S^* \subseteq [n]$, $|S^*| = s$, and $\norm{R_j Y}^2 \ge \norm{R_{j'} Y}^2$, $\forall j \in S^*, \forall j' \in [n] \setminus S^*$.
\end{claim}

\begin{cpf}
This claim follows trivially by writing Problem~$\rst(Y)$ in the form
\begin{align*}
\max_{\substack{S \subseteq [n] \\ |S| = s}} \ 
\sum_{j \in S} \norm{R_j Y}^2.
\end{align*}
\end{cpf}

Claim~\ref{claim multi aux} implies that in order to find an optimal solution to Problem~$\rst(Y)$, it is sufficient to order all values $\norm{R_j Y}^2$, for $j \in [n]$.
Therefore, our next task is to partition all matrices $Y \in \R^{r \times d'}$ based on the order of the values $\norm{R_j Y}^2$, for every $j \in [n]$, that they yield.
Each $\norm{R_j Y}^2$, for $j \in [n]$, is a quadratic polynomial in the entries of $Y$ and every monomial 
is a constant times the product of two variables in the same column of $Y$, i.e., $y_{ki} y_{k'i}$, for $k,k' \in [r]$, $i \in [d']$.
Since we wish to obtain a polyhedral partition, we introduce a new space of variables that allows us to write each $\norm{R_j Y}^2$, for $j \in [n]$, as a linear function.
Formally, we define the space $\E$ that contains one variable for each $y_{ki} y_{k'i}$, for $k,k' \in [r]$, $i \in [d']$.
The dimension of the space $\E$ is therefore $d' \cdot (r^2+r)/2$.
Note that, for each $Y \in \R^{r \times d'}$, there exists a unique corresponding point in $\E$, that we denote by $\ext(Y)$, obtained by computing all the products $y_{ki} y_{k'i}$, for $k,k' \in [r]$, $i \in [d']$.
For each $j \in [n]$, we can now write in time $O(d'r^2)$ a linear function $\ell_j : \E \to \R$ such that $\ell_j (\ext(Y)) = \norm{R_j Y}^2$ for every matrix $Y \in \R^{r \times d'}$.

\begin{claim}
\label{claim multi partition}
There exist a finite index set $T$ of cardinality $O(n^{d'(r^2+r)})$, full-dimensional polyhedra $P^t \subseteq \E$, for $t \in T$, that cover $\E$, and index sets $S^t$, for $t \in T$, with the following property:
For every $t \in T$, and for every $Y$ such that $\ext(Y) \in P^t$, $S^t$ is an optimal solution to Problem~$\rst(Y)$.
The polyhedra $P^t$, for $t \in T$, can be constructed in $O(n^{d'(r^2+r)})$ time.
Furthermore, for each $t \in T$, $S^t$ can be computed in $O(d'n r^2+ n \log n)$ time.
\end{claim}

\begin{cpf}
For every two distinct indices $j,j' \in [n]$,
the hyperplane 
\begin{align}
\label{eq spca hyperplanes}
H_{j,j'} := \{ z \in \E \st \ell_j (z) = \ell_{j'} (z)\}
\end{align}
partitions all points $z \in \E$ based on which of the two values $\ell_j (z)$ and $\ell_{j'} (z)$ is larger.
By considering the hyperplane $H_{j,j'}$ for all distinct pairs of indices $j,j' \in [n]$, we obtain a set $\H$ of $(n^2-n)/2 \le n^2$ hyperplanes in $\E$.
By the hyperplane arrangement theorem, 
the arrangement of $\H$ consists of $O((n^2)^{\dim \E}) = O(n^{d'(r^2+r)})$ full-dimensional polyhedra, and can be constructed in $O(n^{d'(r^2+r)})$ time.
We denote by $P^t$, for $t \in T$, the polyhedra in the arrangement, where $T$ is a finite index set of cardinality $O(n^{d'(r^2+r)})$.
From the definition of the hyperplanes \eqref{eq spca hyperplanes} we have that, if for some $t \in T$ there exists a vector $z^t \in P^t$ that satisfies $\ell_j (z^t) > \ell_{j'} (z^t)$ for two distinct indices $j,j' \in [n]$, then every vector $z \in P^t$ must satisfy $\ell_j (z) \ge \ell_{j'} (z)$.

Next, we explain how the index sets $S^t$, for $t \in T$, are constructed.
To do so, we fix one polyhedron $P^t$, for some $t \in T$, until the end of the proof of the claim.
The hyperplane arrangement theorem also returns explicitly a vector $z^t$ in the interior of $P^t$ \cite{EdeOroSei86}. 
We then compute $\ell_j (z^t)$ for every $j \in [n]$ in time $O(d'n r^2)$.
Since $z^t$ is in the interior of $P^t$, in time $O(n \log n)$ we can find an ordering $j^t_1,j^t_2,\dots,j^t_n$ of the indices $1,\dots,n$ such that
\begin{align*}
\ell_{j^t_1}(z^t)
>
\ell_{j^t_2}(z^t)
> \dots >
\ell_{j^t_n}(z^t).
\end{align*}
From the property of the polyhedra in the arrangement we have that, for every $z$ with $z \in P^t$, 
\begin{align*}
\ell_{j^t_1}(z)
\ge 
\ell_{j^t_2}(z)
\ge \dots \ge 
\ell_{j^t_n}(z).
\end{align*}
In particular, for every $Y$ with $\ext(Y) \in P^t$, we have
\begin{align*}
\ell_{j^t_1}(\ext(Y))
\ge 
\ell_{j^t_2}(\ext(Y))
\ge \dots \ge 
\ell_{j^t_n}(\ext(Y)),
\end{align*}
thus
\begin{align*}
\norm{R_{j^t_1} Y}^2
\ge 
\norm{R_{j^t_2} Y}^2
\ge \dots \ge 
\norm{R_{j^t_n} Y}^2.
\end{align*}
Claim~\ref{claim multi aux} then implies that for each $Y$ such that $\ext(Y) \in P^t$, the set $S^t := \bra{j^t_1,j^t_2,\dots,j^t_s}$ is an optimal solution to Problem~$\rst(Y)$.
\end{cpf}

Let $\S$ be the family of all index sets $S^t$ obtained in Claim~\ref{claim multi partition}, namely
$$
\S := \{ S^t \}_{t \in T}.
$$

\begin{claim}
\label{claim identical final}
The family $\S$ contains an optimal support of Problem~\eqref{prob spca Y}.
\end{claim}

\begin{cpf}
Let $(S^*,Y^*)$ be an optimal solution to Problem~\eqref{prob spca Y}.
Then $S^*$ is an optimal solution to the restricted Problem~$\rst(Y^*)$.
Let $P^t$, for $t \in T$, be a polyhedron such that $\ext(Y^*) \in P^t$, and let $S^t \in \S$ be the corresponding index set.
From Claim~\ref{claim multi partition}, $S^t$ is an optimal solution to Problem~$\rst(Y^*)$.
This implies that the solution $(S^t,Y^*)$ is also optimal to Problem~\eqref{prob spca Y}.
\end{cpf}

Claim~\ref{claim identical final} implies that, in order to find an optimal support of Problem~\eqref{prob spca Y}, it suffices to solve the $|T|$ optimization problems
\begin{align}
\label{prob spca Y inner}
\max_{\substack{Y \in \R^{r \times d'} \\ Y^\transp Y = I_{d'}}} \ 
\Fnorm{R_{S^t} Y}^2
\qquad \qquad \forall t \in T.
\end{align}
In fact, an index set $S^t$, for $t \in T$, which yields the maximum optimal value among Problems~\eqref{prob spca Y inner} is then an optimal support of Problem~\eqref{prob spca Y}.
Each Problem~\eqref{prob spca Y inner} is a \ref{prob pca} problem with an $r \times r$ input matrix.
In particular, the input matrix $R_{S^t}^\transp R_{S^t}$ can be constructed in $O(s r^2)$ time and an optimal solution can be found in $O(r^3)$ time.
This completes the description of the algorithm and the proof of its correctness.

Next, we analyze the total running time of the algorithm presented.
The matrix $R$ is computed in $O(n^3)$ time, 
the linear functions $\ell_j$, for $j \in [n]$, are obtained in $O(d'n r^2)$ time,
the polyhedra $P^t$, for $t \in T$, are constructed  $O(|T|)$ time, 
the sets $S^t$, for $t \in T$, are computed in $O(|T|(d'n r^2 + n \log n))$ time, 
the $|T|$ \ref{prob pca} Problems~\eqref{prob spca Y inner} are solved in $O(|T| (s r^2 + r^3))$ time, and 
the \ref{prob pca}
Problem~\eqref{prob spca final} is solved in $O(s^2r + s^3)$ time.
The total running time is therefore
$$
O\left(
|T| (d'n r^2 + n \log n)
\right)
=
O\left(
n^{d'(r^2+r)} (d'n r^2 + n \log n)
\right).
$$
This concludes the proof of Theorem~\ref{th spca}.
\qed

\section{The maximum-profit integer circulation problem}
\label{sec circulation problem}

In the proof of Theorem~\ref{th spca-ds} we will consider the maximum-profit integer circulation problem.
Hence, before proceeding with the proof, we give a brief overview of this problem and we present optimality conditions and a strongly polynomial-time algorithm to solve it.

Let $D = (V,A)$ be a directed graph.
A vector $f \in \R^A$ is called a \emph{circulation} if $f(\din(v)) = f(\dout(v))$ for each vertex $v \in V$, where $\din(v) = \{wv \in A\}$ and $\dout(v) = \{vw \in A\}$.
A circulation $f$ is said to be \emph{integer} if $f$ has all integer entries.
In a \emph{maximum-profit integer circulation problem} we are given a directed graph $D = (V,A)$,
arc capacities $u \in \Z_+^A$, and arc profits $p \in \Q^A$.
We say that $f$ is a \emph{feasible circulation} if $f$ is an integer circulation in the directed graph $D$ subject to $0 \le f \le u$.
The \emph{profit} of a feasible circulation $f$ is $p^\transp f$.
The goal of the maximum-profit integer circulation problems is that of finding an \emph{optimal circulation,} which is a feasible circulation of maximum profit.
We refer the reader to Chapters 11 and 12 in~\cite{SchBookCO} for a thorough presentation of circulations problems.
We refer the reader to the same book~\cite{SchBookCO} for standard graph theory definitions including that of directed circuit and undirected circuit.

To state the optimality conditions for a maximum-profit integer circulation problem, it will be useful to consider the \emph{residual directed graph} $D_f = (V, A_f)$ of a circulation $f$, where
\begin{align*}
A_f := \{a \st a \in A, \ f_a < u_a\} \cup \{a^\leftarrow \st a \in A, \ f_a > 0\}.
\end{align*}
Here $a^\leftarrow := wu$ if $a = uw$.
For a directed circuit $C$ in $D_f$, we define $\chi^C \in \{0,\pm 1\}^A$ by:
\begin{align*}
\chi^C_a 
:=
\begin{cases}
1 & \text{if $C$ traverses $a$,} \\
-1 & \text{if $C$ traverses $a^\leftarrow$,} \\
0 & \text{if $C$ traverses neither $a$ nor $a^\leftarrow$.}
\end{cases}
\end{align*}
We then define, for every directed circuit $C$ in $D_f$ its \emph{profit} as
\begin{align*}
p(C)
= \sum_{a \in A} \chi_a^C p_a.
\end{align*}

We are now ready to state the optimality conditions, which follow, for example, from Theorem 12.1 in \cite{SchBookCO}. 

\begin{proposition}
\label{prop opt cond}
A feasible circulation $f$ is optimal if and only if each directed circuit in $D_f$ has nonpositive profit.
\end{proposition}

The above optimality conditions are at the basis of Goldberg and Tarjan's strongly polynomial-time algorithm to solve the maximum-profit integer circulation problem~\cite{GolTar88,GolTar89}.
We refer the reader to Section 12.3 in \cite{SchBookCO} for a description of the algorithm.

\begin{proposition}[Corollary 12.2a in \cite{SchBookCO}]
\label{prop GT alg}
An optimal circulation can be found in $O(|V|^2 |A|^3 \log |V|)$ time.
\end{proposition}

%

\section{Proof of Theorem~\ref{th spca-ds}}
\label{sec disjoint}

Consider \ref{prob spca-ds} where the input matrix $K \in \R^{n \times n}$ has rank $r$.
Since the matrix $K$ is positive semidefinite, we can compute an $n \times r$ matrix $R$ such that $K = RR^\transp$ in $O(n^3)$ time, for example using the Cholesky decomposition with complete pivoting.
The objective function of \ref{prob spca-ds} can then be written as $\trace(X^\transp K X) = \Fnorm{R^\transp X}^2 = \sum_{i=1}^d \norm{R^\transp x_i}^2$ and \ref{prob spca-ds} takes the form
\begin{align}
\label{prob spca-ds 2}
\max_{X \in \X} \ 
\sum_{i=1}^d
\norm{R^\transp x_i}^2.
\end{align}

In this proof we use some of the notation introduced in the proof of Theorem~\ref{th spca}.
Namely, for $j \in [n]$, $R_j$ denotes the $j$th row of $R$ and, for $S \subseteq [n]$, $R_S$ denotes the $|S| \times r$ submatrix of $R$ containing only the rows indexed by $S$.

As discussed in Section~\ref{sec intro spca-ds}, the main difficulty in solving Problem~\eqref{prob spca-ds 2} consists in finding an optimal support $\bra{S_i^*}_{i \in [d]}$ of Problem~\eqref{prob spca-ds 2}.
In fact, once $\bra{S_i^*}_{i \in [d]}$ is determined, each optimal vector $x^*_i$, for $i \in [d]$, can be obtained by setting to zero the entries of $x_i^*$ with indices not in $S_i^*$, while the other entries of $x_i^*$ can be obtained by solving the optimization problem 
\begin{align}
\label{prob spca-ds final}
\max_{\substack{x_i \in \R^{\abs{S_i^*}} \\ \norm{x_i} = 1}} \ 
\norm{R_{S_i^*}^\transp x_i}^2.
\end{align}
This is a single-component \ref{prob pca} problem with an input matrix of dimension at most $s \times s$. 
In particular, the input matrix $R_{S_i^*} R_{S_i^*}^\transp$ can be constructed in $O(s^2r)$ time and an optimal solution can be found in $O(s^3)$ time.
Based on this discussion, in the remainder of the proof it suffices to find an optimal support $\bra{S_i^*}_{i \in [d]}$ of Problem~\eqref{prob spca-ds 2}.

The next claim uses Lemma~\ref{lem xy} to replace each vector of variables $x_i \in \R^n$ in Problem~\eqref{prob spca-ds 2} with a vector of variables $y_i \in \R^r$.
In the claim we consider the following two optimization problems:
\begin{align}
\label{prob spca-ds x}
\max_{\substack{S_i \subseteq [n], |S_i| \le s, \ \forall i \in [d] \\ S_i \cap S_{i'} = \emptyset, \ \forall i \neq i' \in [d]}} 
\ \ 
\max_{\substack{x_i \in \R^{\abs{S_i}}, \norm{x_i} = 1, \\ \forall i \in [d]}}
\ 
\sum_{i=1}^d
\norm{(R_{S_i})^\transp x_i}^2,
\end{align}
\begin{align}
\label{prob spca-ds y}
\max_{\substack{S_i \subseteq [n], |S_i| \le s, \ \forall i \in [d] \\ S_i \cap S_{i'} = \emptyset, \ \forall i \neq i' \in [d]}} 
\ \ 
\max_{\substack{y_i \in \R^r, \norm{y_i} = 1, \\ 
\forall i \in [d]}}
\ 
\sum_{i=1}^d
\norm{R_{S_i} y_i}^2.
\end{align}
We say that $\bra{S_i^*}_{i \in [d]}$ is an \emph{optimal support} of Problem~\eqref{prob spca-ds x} if there exist $x_i^*$, for $i \in [d]$, such that $\bra{(S_i^*, x_i^*)}_{i \in [d]}$ is an optimal solution to Problem~\eqref{prob spca-ds x}.
Similarly, we say that $\bra{S_i^*}_{i \in [d]}$ is an \emph{optimal support} of Problem~\eqref{prob spca-ds y} if there exist $y_i^*$, for $i \in [d]$, such that $\bra{(S_i^*, y_i^*)}_{i \in [d]}$ is an optimal solution to Problem~\eqref{prob spca-ds y}.

\begin{claim}
\label{claim disjoint reduction}
The optimal supports of Problems~\eqref{prob spca-ds 2}, \eqref{prob spca-ds x}, \eqref{prob spca-ds y} coincide.
\end{claim}

\begin{cpf}
Lemma~\ref{lem xy}, applied $d$ times with $M := R_{S_i}$, for $i \in [d]$, implies that the optimal supports of Problems~\eqref{prob spca-ds x} and \eqref{prob spca-ds y} coincide. 
Thus we only need to show that the optimal supports of Problems~\eqref{prob spca-ds 2} and \eqref{prob spca-ds x} coincide.
To do so, it suffices to prove the following two statements:
(i) For every feasible solution $\bra{(S_i, x_i)}_{i \in [d]}$ to Problem~\eqref{prob spca-ds x} with objective function value $\gamma$, there is a feasible solution $\bra{\tilde x_i}_{i \in [d]}$ to Problem~\eqref{prob spca-ds 2} with objective function value $\gamma$ such that $\supp(\tilde x_i) \subseteq S_i$ $\forall i \in [d]$;
(ii) For every feasible solution $\bra{\tilde x_i}_{i \in [d]}$ to Problem~\eqref{prob spca-ds 2} with objective function value $\gamma$, there is a feasible solution $\bra{(S_i, x_i)}_{i \in [d]}$ to Problem~\eqref{prob spca-ds x} with objective function value $\gamma$ such that $\supp(\tilde x_i) \subseteq S_i$ $\forall i \in [d]$.

(i).
Let $\bra{(S_i, x_i)}_{i \in [d]}$ be a feasible solution to Problem~\eqref{prob spca-ds x} with objective function value $\gamma$.
For each $i \in [d]$, let $\tilde x_i \in \R^n$ be obtained from $x_i$ by adding zero entries corresponding to the indices not in $S_i$. 
Then $\bra{\tilde x_i}_{i \in [d]}$ is a feasible solution to Problem~\eqref{prob spca-ds 2} with objective function value $\gamma$ such that $\supp(\tilde x_i) \subseteq S_i$ $\forall i \in [d]$.

(ii).
Let $\bra{\tilde x_i}_{i \in [d]}$ be a feasible solution to Problem~\eqref{prob spca-ds 2} with objective function value $\gamma$.
Let $S_i := \supp(\tilde x_i)$, for every $i \in [d]$.
Let $x_i$ be obtained from $\tilde x_i$ by dropping the (zero) entries with indices not in $S_i$.
Then $\bra{(S_i, x_i)}_{i \in [d]}$ is a feasible solution to Problem~\eqref{prob spca-ds x} with objective function value $\gamma$ such that $\supp(\tilde x_i) \subseteq S_i$ $\forall i \in [d]$.
\end{cpf}

Due to Claim \ref{claim disjoint reduction}, in the rest of the proof our goal will be finding an optimal support of Problem~\eqref{prob spca-ds y}.

\subsection{The restricted problem}
\label{sec restricted}

In this section we study the restricted version of Problem~\eqref{prob spca-ds y} obtained by fixing the $d$ vectors of variables $y_i \in \R^r$, for $i \in [d]$.
We denote this restricted problem by $\rst(\bra{y_i}_{i \in [d]})$, and formally define it as 
\begin{align*}
\max_{\substack{S_i \subseteq [n], |S_i| \le s, \ \forall i \in [d] \\ S_i \cap S_{i'} = \emptyset, \ \forall i \neq i' \in [d]}}
\ 
\sum_{i=1}^d
\norm{R_{S_i} y_i}^2.
\end{align*}
Our next goal is to provide a characterization of the optimal solutions to Problem~$\rst(\bra{y_i}_{i \in [d]})$ based on a maximum-profit integer circulation problem.
We refer the reader to Section~\ref{sec circulation problem} for a brief introduction to the maximum-profit integer circulation problem.

\begin{figure}[h]
\centering
\includegraphics[width=0.3\linewidth]{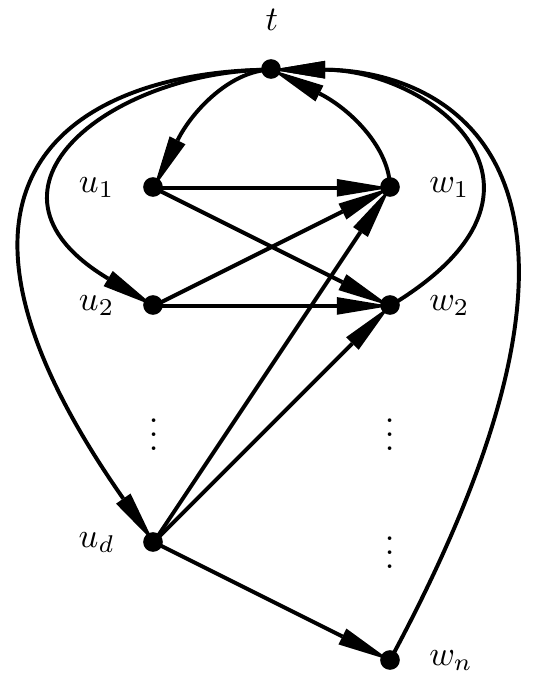}
\caption{The directed graph $D=(V,A)$ considered in Section~\ref{sec restricted}.}
\label{fig digraph}
\end{figure}

In the remainder of the proof, we denote by $D = (V,A)$ the directed graph with vertices $V = U \cup W \cup \{t\}$, where $U=\{u_1,\dots,u_d\}$, $W=\{w_1,\dots,w_n\}$, and with arcs $A = A_0 \cup A_U \cup A_W$, where $A_0 = \{u_i w_j : i \in [d], j \in [n]\}$, $A_U = \{t u_i : i \in [d]\}$, $A_W = \{w_j t : j \in [n]\}$.
The directed graph $D$ is depicted in Figure~\ref{fig digraph}.
We define arc capacities $u \in \Z_+^A$ as $u_a := 1$ if $a \in A_0 \cup A_W$, $u_a := s$ if $a \in A_U$.
We also define arc profits $p \in \Q^A$ by $p_a := \pare{R_j y_i}^2$ if $a = u_i w_j \in A_0$, $p_a := 0$ if $a \in A_U \cup A_W$. 
We then define Problem~$\crc(\bra{y_i}_{i \in [d]})$ as the maximum-profit integer circulation problem on the directed graph $D = (V,A)$, with arc capacities $u$ and arc profits $p$.
We remark that in Problem~$\crc(\bra{y_i}_{i \in [d]})$, only the arc profits depend on $\bra{y_i}_{i \in [d]}$.
The next claim provides a characterization of the optimal solutions to Problem~$\rst(\bra{y_i}_{i \in [d]})$ in terms of optimal circulations to Problem~$\crc(\bra{y_i}_{i \in [d]})$.


\begin{claim}
\label{claim intcir}
Let $\bra{y_i}_{i \in [d]}$ be given.
Then $\bra{S^*_i}_{i \in [d]}$ is an optimal solution to Problem~$\rst(\bra{y_i}_{i \in [d]})$ if and only if $S^*_i := \{j \in [n] : f^*_{u_i w_j} = 1\}$, for $i \in [d]$, where $f^*$ is an optimal circulation to Problem~$\crc(\bra{y_i}_{i \in [d]})$.
\end{claim}

\begin{cpfnodiamond}
To prove the claim, it suffices to prove the following two statements:
(i) For every feasible solution $\bra{S_i}_{i \in [d]}$ to Problem~$\rst(\bra{y_i}_{i \in [d]})$ with objective function value $\gamma$, there is a feasible circulation $f$ to $\crc(\bra{y_i}_{i \in [d]})$ with profit $\gamma$ such that $S_i = \{j \in [n] : f_{u_i w_j} = 1\}$ $\forall i \in [d]$;
(ii) For every feasible circulation $f$ to $\crc(\bra{y_i}_{i \in [d]})$ with profit $\gamma$, the solution $\bra{S_i}_{i \in [d]}$ defined by $S_i := \{j \in [n] : f_{u_i w_j} = 1\}$ $\forall i \in [d]$, is feasible to Problem~$\rst(\bra{y_i}_{i \in [d]})$ and has objective function value $\gamma$.
In the following, we first discuss the mapping between solutions and circulations in (i) and (ii), and then we discuss the correspondence of objective function values and profits in both (i) and (ii).

(i). Let $\bra{S_i}_{i \in [d]}$ be a feasible solution to Problem~$\rst(\bra{y_i}_{i \in [d]})$, i.e.,
$S_i \subseteq [n]$, $|S_i| \le s$, $\forall i \in [d]$, and $S_i \cap S_{i'} = \emptyset$, $\forall i \neq i' \in [d]$.
For every pair $i,j$ such that $j \in S_i$, define $f_{u_i w_j} := 1$, $f_{w_j t} := 1$, and set $f_a := 0$ for every other $a \in A_0 \cup A_W$.
For every $i \in [d]$, define $f_{t u_i} := |S_i|$.
It can be easily checked that $f$ is a feasible circulation to $\crc(\bra{y_i}_{i \in [d]})$ such that $S_i = \{j \in [n] : f_{u_i w_j} = 1\}$ $\forall i \in [d]$.

(ii). Viceversa, let $f$ be a feasible circulation to $\crc(\bra{y_i}_{i \in [d]})$.
Since $u_a = 1$ for every $a \in A_0$, we have $f_a \in \{0,1\}$ for every $a \in A_0$.
For every $i \in [d]$, define $S_i := \{j \in [n] : f_{u_i w_j} = 1\}$.
$u_a = 1$ for every $a \in A_W$ implies that no $j \in [n]$ is in more than one set $S_i$, thus $S_i \cap S_{i'} = \emptyset$, $\forall i \neq i' \in [d]$.
Since $u_a = s$ for every $a \in A_U$, we also have $|S_i| \le s$ for every $i \in [d]$.
Therefore, $\bra{S_i}_{i \in [d]}$ is a feasible solution to Problem~$\rst(\bra{y_i}_{i \in [d]})$.

The claim follows since objective function values and profits coincide in both mappings (i) and (ii):
\begin{align*}
p^\transp f
=
\sum_{i=1}^d
\sum_{j =1}^n
p_{u_i w_j}
f_{u_i w_j}
=
\sum_{i=1}^d
\sum_{j \in S_i}
p_{u_i w_j}
=
\sum_{i=1}^d
\sum_{j \in S_i}
\pare{R_j y_i}^2
=
\sum_{i=1}^d
\norm{R_{S_i} y_i}^2.
\tag*{$\diamond$}
\end{align*}
\end{cpfnodiamond}

\subsection{A polynomial arrangement}
\label{sec arrangement}

Claim~\ref{claim intcir} implies that in order to find an optimal solution to Problem~$\rst(\bra{y_i}_{i \in [d]})$, it is sufficient to find an optimal circulation to Problem~$\crc(\bra{y_i}_{i \in [d]})$.
Thus we now focus on the latter problem.
The optimality conditions stated in Proposition~\ref{prop opt cond} imply that in order to understand an optimal circulation to Problem~$\crc(\bra{y_i}_{i \in [d]})$, it is important to understand the sign of the profits of all directed circuits in $D_f$, for any feasible circulation $f$.
Note that any directed circuit $C$ in $D_f$, for a feasible circulation $f$, gives an undirected circuit $C'$ in $D$.
For an undirected circuit $C'$ in $D$, we define $\chi^{C'} \in \{0,\pm 1\}^A$ by:
\begin{align*}
\chi^{C'}_a 
:=
\begin{cases}
1 & \text{if $C'$ traverses $a$ forward,} \\
-1 & \text{if $C'$ traverses $a$ backward,} \\
0 & \text{if $C'$ does not traverse $a$.}
\end{cases}
\end{align*}
We then define, for every undirected circuit $C'$ in $D$, its \emph{profit} as
\begin{align*}
p(C') = \sum_{a \in A} \chi_a^{C'} p_a.
\end{align*}
In this way we obtain that, if a directed circuit $C$ in $D_f$, for some feasible circulation $f$, gives the undirected circuit $C'$ in $D$, then we have $p(C) = p(C')$.
From the above discussion, in order to understand the sign of the profits of all directed circuits in $D_f$, for any feasible circulation $f$, it suffices to understand the signs of the profits of all undirected circuits in $D$.
From now on, we denote by $\C$ the set of undirected circuits in $D$.
The structure of the directed graph $D$ implies that each undirected circuit in $\C$ can contain at most $d$ vertices in $W$.
Thus we obtain $|\C| = O((dn)^d)$.

Our next task is to partition the $dr$-dimensional space of all $d$ vectors $\bra{y_i}_{i \in [d]}$, where each $y_i$ is in $\R^r$, based on the sign of the values $p(C')$, for every $C' \in \C$, that they yield.
Each $p(C')$, for $C' \in \C$, can be written as a linear function of arc profits
\begin{align*}
p(C')
= \sum_{a \in A} \chi_a^{C'} p_a
= \sum_{u_iw_j \in A_0} \chi_{u_iw_j}^{C'} p_{u_iw_j}.
\end{align*}
Each arc profit $p_{u_iw_j} = \pare{R_j y_i}^2$, for $i \in [d]$, $j \in [n]$, is a quadratic polynomial in the entries of the vector $y_i$, and 
every monomial is a constant times the product of two variables in the 
vector $y_i$, i.e., $(y_i)_k (y_i)_{k'}$, for 
$k,k' \in [r]$.
Since we wish to obtain a polyhedral partition, we introduce a new space of variables that allows us to write each $p(C')$, for $C' \in \C$, as a linear function. 
Formally, we define the space $\E$ that contains one variable for each $(y_i)_k (y_i)_{k'}$, for $i \in [d]$, $k,k' \in [r]$.
The dimension of the space $\E$ is therefore $d \cdot (r^2+r) / 2$.
Note that, for every $d$ vectors $\bra{y_i}_{i \in [d]}$, where each $y_i$ is in $\R^r$, there exists a unique corresponding point in $\E$, that we denote by $\ext(\bra{y_i}_{i \in [d]})$, obtained by computing all the products $(y_i)_k (y_i)_{k'}$, for $i \in [d]$, $k,k' \in [r]$.
For each arc $u_iw_j$, $i \in [d]$, $j \in [n]$, we can now write in time $O(r^2)$ a linear function $\ell_{u_iw_j} : \E \to \R$ such that $\ell_{u_iw_j} (\ext(\bra{y_i}_{i \in [d]})) = \pare{R_j y_i}^2$ for every $\bra{y_i}_{i \in [d]}$.
As a consequence, for each $C' \in \C$, we can write a linear function $\ell_{C'} : \E \to \R$ such that $\ell_{C'} (\ext(\bra{y_i}_{i \in [d]})) = p(C')$ for every $\bra{y_i}_{i \in [d]}$.
Note that all these linear functions can be constructed in time $O(dnr^2 + dr^2|\C|)=O(dnr^2 + d^{d+1}r^2n^d)$.

\begin{claim}
\label{claim disjoint partition}
There exist a finite index set $T$ of cardinality $O((dn)^{d^2(r^2+r)/2})$,
full-dimensional polyhedra $P^t \subseteq \E$, for $t \in T$, that cover $\E$, and index sets $\bra{S^t_i}_{i \in [d]}$,
for $t \in T$, 
with the following property:
For every $t \in T$, and for every $\bra{y_i}_{i \in [d]}$ such that $\ext(\bra{y_i}_{i \in [d]}) \in P^t$, $\bra{S^t_i}_{i \in [d]}$ is an optimal solution to Problem~$\rst(\bra{y_i}_{i \in [d]})$.
The polyhedra $P^t$, for $t \in T$, can be constructed in $O((dn)^{d^2(r^2+r)/2}))$ time.
Furthermore, for each $t \in T$, $\bra{S^t_i}_{i \in [d]}$ can be computed in $O(dn r^2 + 
d^3 n^5 \log n)$ time.
\end{claim}

\begin{cpf}
%
For every $C' \in \C$, the hyperplane
\begin{align}
\label{eq spca-ds hyperplanes b}
H_{C'} := \{ z \in \E \st \ell_{C'} (z) = 0\}
\end{align}
partitions all points $z \in \E$ based on the sign of $\ell_{C'} (z)$.
%
By considering the hyperplane $H_{C'}$ for all $C' \in \C$, 
we obtain a set $\H$ of $|\C| = O((dn)^{d})$ hyperplanes in $\E$.
By the hyperplane arrangement theorem, the arrangement of $\H$ consists of $O((dn)^{d \cdot \dim \E}) = O((dn)^{d^2(r^2+r)/2})$ full-dimensional polyhedra, and can be constructed in $O((dn)^{d^2(r^2+r)/2})$ time.
We denote by $P^t$, for $t \in T$, the polyhedra in the arrangement, where $T$ is a finite index set of cardinality $O((dn)^{d^2(r^2+r)/2})$.
From the definition of the hyperplanes \eqref{eq spca-ds hyperplanes b} we have that, if for some $t \in T$ there exists a vector $z^t \in P^t$ that satisfies $\ell_{C'} (z^t) < 0$ for some $C' \in \C$, then every vector $z \in P^t$ must satisfy $\ell_{C'} (z) \le 0$.

Next, we explain how the index sets $\bra{S^t_i}_{i \in [d]}$, for $t \in T$, are constructed. 
To do so, we fix one polyhedron $P^t$, for some $t \in T$, until the end of the proof of the claim.
Due to Claim~\ref{claim intcir}, it suffices to show that we can construct a circulation $f$ that is an optimal circulation to every Problem~$\crc(\bra{y_i}_{i \in [d]})$ for all $\bra{y_i}_{i \in [d]}$ with $\ext(\bra{y_i}_{i \in [d]}) \in P^t$.
To obtain this optimal circulation we will use a vector $z^t$ in the interior of $P^t$, which is returned explicitly by the hyperplane arrangement theorem \cite{EdeOroSei86}. 
Then, we define Problem~$\crc(z^t)$ as the problem obtained from Problem~$\crc(\bra{y_i}_{i \in [d]})$ for any $\bra{y_i}_{i \in [d]}$ with $\ext(\bra{y_i}_{i \in [d]}) \in P^t$, by replacing the arc profits with the one induced by $z^t$.
Precisely, Problem~$\crc(z^t)$ is the maximum-profit integer circulation problem on the directed graph $D = (V,A)$ defined in Section~\ref{sec restricted}, with arc capacities $u \in \Z_+^A$ defined in Section~\ref{sec restricted}, and arc profits $p^t \in \Q^A$ defined by $p^t_a := \ell_{u_iw_j} (z^t)$ if $a = u_i w_j \in A_0$, $p^t_a := 0$ if $a \in A_U \cup A_W$.  
Note that these arc profits can be computed in time $O(dn r^2)$.

From Proposition~\ref{prop GT alg}, an optimal circulation $f^*$ to Problem~$\crc(z^t)$ can be found in $O(|V|^2 |A|^3 \log |V|)$ time.
Since $|V| = O(n)$ and $|A| = O(dn)$, we can obtain $f^*$ in $O(d^3 n^5 \log n)$ time.
We now show that $f^*$ is an optimal circulation to every Problem~$\crc(\bra{y_i}_{i \in [d]})$ for all $\bra{y_i}_{i \in [d]}$ with $\ext(\bra{y_i}_{i \in [d]}) \in P^t$.
So we fix an arbitrary $\bra{\bar y_i}_{i \in [d]}$ with $\ext(\bra{\bar y_i}_{i \in [d]}) \in P^t$.
In the remainder of the proof we will denote by $p^t$ the profits in Problem~$\crc(z^t)$ and by $\bar p$ the profits in Problem~$\crc(\bra{\bar y_i}_{i \in [d]})$.
Since $f^*$ is a feasible circulation to Problem~$\crc(z^t)$, it is also a feasible circulation to Problem~$\crc(\bra{\bar y_i}_{i \in [d]})$.
This is because the two problems share the same directed graph and the same arc capacities.
Furthermore, the residual directed graph $D_{f^*}$ is the same in both problems.
From the optimality conditions stated in Proposition~\ref{prop opt cond}, we know that $p^t(C) \le 0$ for every directed circuit $C$ in $D_{f^*}$.
From the definition of the hyperplanes \eqref{eq spca-ds hyperplanes b} and the fact that $z^t$ is in the interior of $P^t$, we obtain that $p^t(C) < 0$ for every directed circuit $C$ in $D_{f^*}$.
Since $\ext(\bra{\bar y_i}_{i \in [d]}) \in P^t$, we then have $\bar p(C) \le 0$ for every directed circuit $C$ in $D_{f^*}$.
Again from the optimality conditions in Proposition~\ref{prop opt cond}, we obtain that $f^*$ is an optimal circulation to Problem~$\crc(\bra{\bar y_i}_{i \in [d]})$.
We have thereby shown that $f^*$ is an optimal circulation to every Problem~$\crc(\bra{y_i}_{i \in [d]})$ for all $\bra{y_i}_{i \in [d]}$ with $\ext(\bra{y_i}_{i \in [d]}) \in P^t$.
An optimal solution to all Problems~$\rst(\bra{y_i}_{i \in [d]})$ for all $\bra{y_i}_{i \in [d]}$ with $\ext(\bra{y_i}_{i \in [d]}) \in P^t$ can then be obtained as described in Claim~\ref{claim intcir}.
The total running time to compute $\bra{S^t_i}_{i \in [d]}$ is
$
O( 
dn r^2 + 
d^3 n^5 \log n
)
$
\end{cpf}

Let $\S$ be the family of all index sets $\bra{S^t_i}_{i \in [d]}$ obtained in Claim~\ref{claim disjoint partition}, namely
$$
\S := \{ \bra{S^t_i}_{i \in [d]} \}_{t \in T}.
$$

\begin{claim}
\label{claim disjoint final}
The family $\S$ contains an optimal support of Problem~\eqref{prob spca-ds y}.
\end{claim}

\begin{cpf}
Let $\bra{(S_i^*, y_i^*)}_{i \in [d]}$ be an optimal solution to Problem~\eqref{prob spca-ds y}.
Then $\bra{S_i^*}_{i \in [d]}$ is an optimal solution to the restricted Problem~$\rst(\bra{y_i^*}_{i \in [d]})$.
Let $P^t$, for $t \in T$, be a polyhedron such that $\ext(\bra{y_i^*}_{i \in [d]}) \in P^t$, and let $\bra{S_i^t}_{i \in [d]} \in \S$ be the corresponding index sets.
From Claim~\ref{claim disjoint partition}, $\bra{S_i^t}_{i \in [d]}$ is an optimal solution to Problem~$\rst(\bra{y_i^*}_{i \in [d]})$.
This implies that the solution $\bra{S_i^t, y_i^*}_{i \in [d]}$ is also optimal to Problem~\eqref{prob spca-ds y}.
\end{cpf}

Claim~\ref{claim disjoint final} implies that, in order to find an optimal support of Problem~\eqref{prob spca-ds y}, it suffices to solve the $|T|$ optimization problems
\begin{align}
\label{prob spca-ds y inner}
\max_{\substack{y_i \in \R^r, \norm{y_i} = 1, \\ 
\forall i \in [d]}}
\ 
\sum_{i=1}^d
\norm{R_{S^t_i} y_i}^2
\qquad \qquad \forall t \in T.
\end{align}
In fact, a $\bra{S_i^t}_{i \in [d]}$, for $t \in T$, which yields the maximum optimal value among Problems~\eqref{prob spca-ds y inner} is then an optimal support of Problem~\eqref{prob spca-ds y}.
Each Problem~\eqref{prob spca-ds y inner} can be decomposed into the $d$ optimization problems
\begin{align}
\label{prob spca-ds y inner single}
\max_{\substack{y_i \in \R^r \\ \norm{y_i} = 1}} \ 
\norm{R_{S^t_i} y_i}^2
\qquad \qquad \forall i \in [d].
\end{align}
Each Problem~\eqref{prob spca-ds y inner single} is a single-component \ref{prob pca} problem with an $r \times r$ input matrix.
In particular, the input matrix $R_{S^t_i}^\transp R_{S^t_i}$ can be constructed in $O(s r^2)$ time and an optimal solution can be found in $O(r^3)$ time.
This completes the description of the algorithm and the proof of its correctness.

Next, we analyze the total running time of the algorithm presented.
The matrix $R$ is computed in $O(n^3)$ time, 
the linear functions $\ell_{C'}$, for $C' \in \C$, are constructed in $O(dnr^2 + d^{d+1}r^2n^d)$ time,
the polyhedra $P^t$, for $t \in T$, are constructed  $O(|T|)$ time, 
the sets $\bra{S^t_i}_{i \in [d]}$, for $t \in T$, are computed in $O(|T|(dn r^2 + d^3 n^5 \log n))$ time,
the $|T| d$ \ref{prob pca} Problems~\eqref{prob spca-ds y inner single} are solved in $O(|T| d (s r^2 + r^3))$ time, 
and the $d$ \ref{prob pca} Problems~\eqref{prob spca-ds final} are solved in $O(d(s^2r + s^3))$ time.
The total running time is therefore 
$$
O\left( 
|T|(dn r^2 + d^3 n^5 \log n)
\right)
=
O\left( 
(dn)^{d^2(r^2+r)/2} (dn r^2 + d^3 n^5 \log n)
\right).
$$
This concludes the proof of Theorem~\ref{th spca-ds}.
\qed


\ifthenelse {\boolean{MPA}}
{
\bibliographystyle{spmpsci}
}
{
\bibliographystyle{plain}
}


\begin{thebibliography}{10}
\providecommand{\url}[1]{{#1}}
\providecommand{\urlprefix}{URL }
\expandafter\ifx\csname urlstyle\endcsname\relax
  \providecommand{\doi}[1]{DOI~\discretionary{}{}{}#1}\else
  \providecommand{\doi}{DOI~\discretionary{}{}{}\begingroup
  \urlstyle{rm}\Url}\fi

\bibitem{AstPapKar11}
Asteris, M., Papailiopoulos, D., Karystinos, G.: Sparse principal component of
  a rank-deficient matrix.
\newblock In: Proceedings of ISIT (2011)

\bibitem{AstPapKar14}
Asteris, M., Papailiopoulos, D., Karystinos, G.: The sparse principal component
  of a constant-rank matrix.
\newblock IEEE Transactions on Information Theory pp. 2281--2290 (2014)

\bibitem{AstPapKyrDim15}
Asteris, M., Papailiopoulos, D., Kyrillidis, A., Dimakis, A.: Sparse {PCA} via
  bipartite matchings.
\newblock In: Proceedings of NIPS (2015)

\bibitem{BerBer19}
Berk, L., Bertsimas, D.: Certifiably optimal sparse principal component
  analysis.
\newblock Mathematical Programming Computation \textbf{11}, 381--420 (2019)

\bibitem{BerTsi97}
Bertsimas, D., Tsitsiklis, J.: Introduction to Linear Optimization.
\newblock Athena Scientific, Belmont, MA (1997)

\bibitem{BouDriMag11}
Boutsidis, C., Drineas, P., Magdon-Ismail, M.: Sparse features for {PCA}-like
  linear regression.
\newblock In: Proceedings of NIPS, pp. 2285--2293 (2011)

\bibitem{CadJol95}
Cadima, J., Jolliffe, I.: Loading and correlations in the interpretation of
  principle compenents.
\newblock Journal of Applied Statistics \textbf{22}(2), 203--214 (1995)

\bibitem{ChaPapRub16}
Chan, S., Papailiopoulos, D., Rubinstein, A.: On the worst-case approximability
  of sparse {PCA}.
\newblock Proceedings of COLT  (2016)

\bibitem{dAsBacGha08}
d'Aspremont, A., Bach, F., Ghaoui, L.: Optimal solutions for sparse principal
  component analysis.
\newblock The Journal of Machine Learning Research \textbf{9}, 1269--1294
  (2008)

\bibitem{dAsBacGha14}
d'Aspremont, A., Bach, F., Ghaoui, L.: Approximation bounds for sparse
  principal component analysis.
\newblock Mathematical Programming, Series B pp. 89--110 (2014)

\bibitem{dAsElGJorLan07}
d'Aspremont, A., El~Ghaoui, L., Jordan, M., Lanckriet, G.: A direct formulation
  for sparse {PCA} using semidefinite programming.
\newblock SIAM review \textbf{49}(3), 434--448 (2007)

\bibitem{DeyMazWan18}
Dey, S., Mazumder, R., Wang, G.: A convex integer programming approach for
  optimal sparse {PCA}.
\newblock arXiv preprint arXiv:1810.09062  (2018)

\bibitem{EdeOroSei86}
Edelsbrunner, H., O'Rourke, J., Seidel, R.: Constructing arrangements of lines
  and hyperplanes with applications.
\newblock SIAM Journal on Computing \textbf{15}(2), 341--363 (1986)

\bibitem{GolTar88}
Goldberg, A., Tarjan, R.: Finding minimum-cost circulations by canceling
  negative cycles.
\newblock In: Proceedings of STOC, pp. 388--397 (1988)

\bibitem{GolTar89}
Goldberg, A., Tarjan, R.: Finding minimum-cost circulations by canceling
  negative cycles.
\newblock Journal of the Association for Computing Machinery \textbf{36},
  873--886 (1989)

\bibitem{HasTibWai15}
Hastie, T., Tibshirani, R., Wainwright, M.: Statistical learning with sparsity.
\newblock CRC press (2015)

\bibitem{HeMonPar10}
He, Y., Monteiro, R., Park, H.: An efficient algorithm for rank-1 sparse {PCA}.
\newblock working paper  (2010)

\bibitem{JolTreUdd03}
Jolliffe, I., Trendafilov, N., Uddin, M.: A modified principal component
  technique based on the lasso.
\newblock Journal of Computational and Graphical Statistics \textbf{12}(3),
  531--547 (2003)

\bibitem{JouNesRicSep10}
Journ\'ee, M., Nesterov, Y., Richt\'arik, P., Sepulchre, R.: Generalized power
  method for sparse principal component analysis.
\newblock The Journal of Machine Learning Research \textbf{11}, 517--553 (2010)

\bibitem{KarLia10}
Karystinos, G., Liavas, A.: Efficient computation of the binary vector that
  maximizes a rank-deficient quadratic form.
\newblock IEEE Transactions on Information Theory \textbf{56}(7), 3581--3593
  (2010)

\bibitem{KarPad07}
Karystinos, G., Pados, D.: Rank-2-optimal adaptive design of binary spreading
  codes.
\newblock IEEE Transactions on Information Theory \textbf{53}(9), 3075--3080
  (2007)

\bibitem{Mac94}
Mackenthun, K.: A fast algorithm for multiple-symbol differential detection of
  {MPSK}.
\newblock IEEE Transactions on Communications \textbf{42}(2/3/4), 1471--1474
  (1994)

\bibitem{Mac09}
Mackey, L.: Deflation methods for sparse {PCA}.
\newblock In: Proceedings of NIPS, vol.~21, pp. 1017--1024 (2009)

\bibitem{Mag17}
Magdon-Ismail, M.: {NP}-hardness and inapproximability of sparse {PCA}.
\newblock Information Processing Letters pp. 35--38 (2017)

\bibitem{MogWeiAvi06NIPS}
Moghaddam, B., Weiss, Y., Avidan, S.: Spectral bounds for sparse {PCA}: Exact
  and greedy algorithms.
\newblock In: Proceedings of NIPS, vol.~18, p. 915 (2006)

\bibitem{MotKriAna07}
Motedayen, I., Krishnamoorthy, A., Anastasopoulos, A.: Optimal joint
  detection/estimation in fading channels with polynomial complexity.
\newblock IEEE Transactions on Information Theory \textbf{53}(1), 209--223
  (2007)

\bibitem{PapDimKor13}
Papailiopoulos, D., Dimakis, A., Korokythakis, S.: Sparse {PCA} through
  low-rank approximations.
\newblock In: Proceedings of ICML (2013)

\bibitem{SchBookCO}
Schrijver, A.: Combinatorial Optimization. Polyhedra and Efficiency.
\newblock Springer-Verlag, Berlin (2003)

\bibitem{ShaBenMLbook}
Shalev-Shwartz, S., Ben-David, S.: Understanding Machine Learning.
\newblock Cambridge University Press (2014)

\bibitem{SigBuh08}
Sigg, C., Buhmann, J.: Expectation-maximization for sparse and non-negative
  {PCA}.
\newblock In: Proceedings of ICML, pp. 960--967 (2008)

\bibitem{Swe01}
Sweldens, W.: Fast block noncoherent decoding.
\newblock IEEE Communications Letters \textbf{5}(4), 132--134 (2001)

\bibitem{VuLei12}
Vu, V., Lei, J.: Minimax rates of estimation for sparse {PCA} in high
  dimensions.
\newblock In: Proceedings of AIStats, pp. 1278--1286 (2012)

\bibitem{YuaZha13}
Yuan, X., Zhang, T.: Truncated power method for sparse eigenvalue problems.
\newblock Journal of Machine Learning Research \textbf{14}, 899--925 (2013)

\bibitem{ZhadAsGha12}
Zhang, Y., d'Aspremont, A., L., G.: Sparse {PCA}: Convex relaxations,
  algorithms and applications.
\newblock In: Handbook on Semidefinite, Conic and Polynomial Optimization, pp.
  915--940. Springer (2012)

\bibitem{ZouHasTib06}
Zou, H., Hastie, T., Tibshirani, R.: Sparse principal component analysis.
\newblock Journal of computational and graphical statistics \textbf{15}(2),
  265--286 (2006)

\end{thebibliography}

\end{document}